\documentclass[lettersize]{article}

\usepackage{PRIMEarxiv}

\usepackage[utf8]{inputenc} % allow utf-8 input
\usepackage[T1]{fontenc}    % use 8-bit T1 fonts
\usepackage{hyperref}       % hyperlinks
\usepackage{url}            % simple URL typesetting
\usepackage{booktabs}       % professional-quality tables
\usepackage{amsfonts}       % blackboard math symbols
\usepackage{nicefrac}       % compact symbols for 1/2, etc.
\usepackage{microtype}      % microtypography
\usepackage{fancyhdr}       % header
\usepackage{graphicx}       % graphics
\graphicspath{{media/}}     % organize your images and other figures under media/ folder
\usepackage{amsmath}
\usepackage{graphicx}
\usepackage{pdflscape}
\usepackage{rotating} % Rotating table
\usepackage{lmodern}
\usepackage{pdflscape}
\usepackage{lmodern}
\usepackage{array, multirow}

\usepackage{afterpage}
\usepackage{capt-of}% or use the larger `caption` package
\usepackage{multirow}
\usepackage{color}

\title{ESTIMATE DEFORMATION CAPACITY OF NON-DUCTILE RC SHEAR WALLS USING EXPLAINABLE BOOSTING MACHINE}
\author{Zeynep Tuna Deger (1*), Gulsen Taskin Kaya (1), John W. Wallace (2) \\
  (1) Istanbul Technical University, (2) University of California, Los Angeles \\
  (*) corresponding author \\
  \texttt{\{zeynep.tuna@itu.edu.tr, gulsen.taskin@itu.edu.tr, wallacej@ucla.edu\}} \\
}

\begin{document}
\maketitle

\begin{abstract}
Machine learning is becoming increasingly prevalent for tackling challenges in earthquake engineering and providing fairly reliable and accurate predictions. However, it is mostly unclear how decisions are made because machine learning models are generally highly sophisticated, resulting in opaque black-box models. Machine learning models that are naturally interpretable and provide their own decision explanation, rather than using an explanatory, are more accurate in determining what the model actually computes. With this motivation, this study aims to develop a fully explainable machine learning model to predict the deformation capacity of non-ductile reinforced concrete shear walls based on experimental data collected worldwide. The proposed Explainable Boosting Machines (EBM)-based model is an interpretable, robust, naturally explainable glass-box model, yet provides high accuracy comparable to its black-box counterparts. The model enables the user to observe the relationship between the wall properties and the deformation capacity by quantifying the individual contribution of each wall property as well as the correlations among them. The mean coefficient of determination $R^2$ and the mean ratio of predicted to actual value based on the test dataset are 0.92 and 1.05, respectively.  The proposed predictive model stands out with its overall consistency with scientific knowledge, practicality, and interpretability without sacrificing high accuracy. 
\end{abstract}


\section{Introduction}
\label{sec:sample1}

Shear walls are typically utilized as the primary elements to resist lateral loads in reinforced concrete buildings. Towards capacity design assumptions, shear walls are designed to exhibit ductile behavior by providing adequate reinforcement and proper detailing. However, experimental studies have shown that walls with an aspect ratio smaller than $1.5$ (i.e., squat walls) and those with poor reinforcement and detailing, despite their higher aspect ratio, end up showing brittle failure (e.g., diagonal tension, web crushing) \cite{american2017asce,massone2004load, sittipunt2001cyclic}. Such walls are often observed in buildings not designed according to modern seismic codes and are prone to severe damage \cite{wallace2012damage, fema454}. As the performance-based design and assessment approach has gained importance concordant with hazard mitigation efforts, there has been an increasing need and demand for reliable models to predict structural behavior under seismic actions. This objective is particularly important for walls that exhibit shear behavior as the nonlinear deformation capacity of such walls is assumed to be zero, potentially leading to technical and economical over-conservation. More realistic solutions can be achieved if their behavior is accurately estimated and considered in seismic performance evaluation.

The prediction of structural behavior has been achieved through the use of predictive equations or models that are developed based on available experimental data. Recently, machine learning (ML) methods have gained significant attention structural/earthquake engineering field and have demonstrated promising results despite the scarcity of data (compared to much larger data available in fields such as computer vision and image processing). Black-box models, with their high complexity and nonlinearity, often represent the input-output relationship better than the interpretable models in classification and regression applications. However, they are not necessarily consistent with true physical behavior. There are examples of misleading conclusions of black-box models in scientific and engineering applications \cite{lazer2014parable,douglas2008survey,karpatne2022knowledge}. Therefore, despite the high accuracy they achieve, black-box models are not completely accepted in earthquake engineering society. To leverage the advantages of the developments in artificial intelligence without ignoring the physical behavior, there has been recent research efforts that incorporates black-box machine learning methods with physical knowledge \cite{karpatne2022knowledge,LUO2022101568,ZHOU2022101642,aladsani2022explainable}. This study takes this issue a step further and integrates an \textit{explainable} machine learning approach (versus black-box) with existing physics-based understanding of seismic behavior to estimate the deformation capacity of non-ductile shear walls.

\section{Literature Review}
Research efforts in the literature to estimate wall deformation capacity have produced empirical models, some recently adopted by building codes \cite{abdullah2019drift}; however, they are relatively limited compared to other behavior features such as shear strength or failure mode. Earlier models were mainly developed using a limited number of experimental results \cite{paulay1982ductility,kazaz2012deformation} or were trained using a single dataset; that is, they were not trained and tested based on unmixed data \cite{abdullah2019drift,grammatikou2015strength}. Over time, as machine learning is embraced in the earthquake engineering field \cite{mangalathu2020data,feng2021interpretable,zhang2022prediction,deger2022novel,deger2022glass,aladsani2022explainable} and new experiments are conducted, more advanced models have been developed. Yet, two main issues are encountered: (i) Some models used simple approaches such as linear regression for the sake of interpretability \cite{deger2019empirical} and sacrificed overall accuracy (or had large dispersion). One might think that accurate models that predict relatively complicated behavior attributes can only be achieved by increasing model complexity; however, literature studies have shown that this may cause problems with the structure and distribution of the data \cite{johnson2019survey,kailkhura2019reliable}. More importantly, urging the model to develop complex relationships to achieve higher performance typically leads to black-box models where internal mechanisms include highly nonlinear, complex relations. (ii) Such black-box models achieve high overall accuracy at the cost of explainability \cite{zhang2022prediction}. Researchers that acknowledge the significance of interpretability employed model-agnostic, local or global explanation methods (e.g., SHapley Additive exPlanation, Local Interpretable Model-agnostic Explanations) to interpret the decision mechanism of their models \cite{aladsani2022explainable}. Such algorithms are not fully verified \cite{kumar2020problems,rudin2019stop}; besides, they are approximate approaches. Moreover,  despite their broadening use and high accuracy, the black-box models are not entirely accepted in the earthquake engineering society as their internal relations are opaque and, in some cases, not entirely reliable \cite{molnar2020interpretable}. Therefore, it is critical to understand how the model makes the decision/estimation to (i) verify that the model is physically meaningful, (ii) develop confidence in the predictive model, and (iii) broaden existing scientific knowledge with new insights.This study addresses this need and fills this important research gap by using domain-specific knowledge to evaluate and validate the decisions made by ML methods. Unlike the existing ML-based predictive models (\cite{aladsani2022explainable,zhang2022prediction}), the proposed model aims particularly at the deformation capacity of non-ductile shear walls and is naturally transparent and interpretable.

Concerns regarding the trustworthiness and transparency of the black-box models motivated the development of a relatively new research area known as explainable artificial intelligence (XAI) \cite{adadi2018peeking,lipton2018mythos}. The XAI aims to provide a set of machine learning (ML) techniques for building more comprehensible and understandable models while maintaining a high level of learning performance. The strategies used in XAI are divided into two main categories: explaining existing black-box models (post-hoc explainability) and generating  glass-box (transparent) models. In the former, interpretability is confined to the usage of certain so-called explanatory algorithms that are employed to explain a black-box model,  while in the latter, a predictive model is fully comprehensible and interpretable by humans.  A model should have certain qualities to be considered a transparent model such as decomposability, algorithmic transparency, and simulatability \cite{arrieta2020explainable}. The  decomposability relates to the ability to explain each model component in terms of the inputs' contributions or correlations, whereas simulatability refers to the number of parameters (input) in the model representation (the less is the more understandable). The algorithmically transparent models enable a clear comprehension of the model' behavior for predicting any given output from its input data. %The transparent  models are highly needed approaches in the fields where decisions are critical, but their number is relatively limited compared to black-box models \cite{doshi2017towards}. The primary reason is that black-box models can make considerably better decisions than transparent models. This is because highly complicated functions are employed in the training of black-box machine learning models, allowing the nonlinear structure hidden in the data to be captured. Although post-hoc explainability is the most often utilized method in the literature, transparent models can be effectively substituted by complex models without sacrificing the performance \cite{selbst2018intuitive}.
%The purpose of black-box models is to optimize performance by ignoring the level of complexity of the functions. In contrast, 
%The transparent (glass-box)  models aim to construct learning models with  simpler functions to maximize performance as well as to enable humans to understand and comprehend the learning model. However, due to the utilization of simple functions during the learning phase, performance levels are typically low \cite{linardatos2020explainable}. The structure of the functions employed in the construction of the predictive model can be complicated to boost the performance, but this may reduce the level of explainability. 
Therefore, transparent  models are highly needed approaches in fields where decisions are critical, but their performances are typically very low. Machine learning models that can maintain the tradeoff between performance and explainability, i.e., converging to the performance of black-box models while still providing explainability, would significantly address the demands in earthquake engineering society. In this context, explainable boosting machine (EBM) \cite{nori2019interpretml}, a recently developed method belonging to the family of Generalized Additive Models (GAMs) \cite{hastie2009additive}, is a highly accurate and transparent ML method delivering an explicit and fully explainable predictive model. The EBM has been utilized in the literature to solve a  variety of problems, including detecting common flaws in data \cite{chen2021using}, diagnosing COVID-19 using blood test variables \cite{thimoteo2022explainable}, predicting diseases such as Alzheimer \cite{sarica2021explainable}, or Parkinson \cite{sarica2022explainable}, and has shown to outperform black-box models with the additional benefit of being an inherently explainable predictive model. 

%\textcolor{blue}{The literature review mentioned above enhances the accuracy of deformation capacity estimations for shear walls; however, none of the models particularly aim at non-ductile shear walls. Besides, current predictive models utilize black-box models that address the interpretability issue using model-agnostic explanation methods. This study fills an important research gap by developing a fully explainable and accurate predictive model for estimating the deformation capacity of non-ductile shear walls.}

In this study, the EBM is used for the first time in the earthquake engineering field to construct an EBM-based predictive model for estimating deformation capacity on non-ductile RC shear walls.  The inputs of the predictive model are designated as the shear wall design properties (e.g., wall geometry, reinforcing ratio), whereas the output is one of the constitutive components of the nonlinear wall behavior, that is, the deformation capacity. The main contributions of this research are highlighted as follows:
\begin{itemize}
    \item A fully transparent and interpretable predictive model is developed to estimate the deformation capacity of RC shear walls that are failed in pure shear or shear-flexure interaction. 
    \item The proposed model meets all desired properties, i.e., decomposability, algorithmic transparency, and simulatability, without compromising high performance. 
    \item This study integrates novel computational methods (i.e., EBM) and domain-based knowledge to formalize complex engineering knowledge. The proposed model's overall consistency with a physics-based understanding of seismic behavior is verified. 
\end{itemize}

\section{The RC Shear Wall Database}
\begin{figure*}[!h]
	\centering
	\includegraphics[width=5.2in]{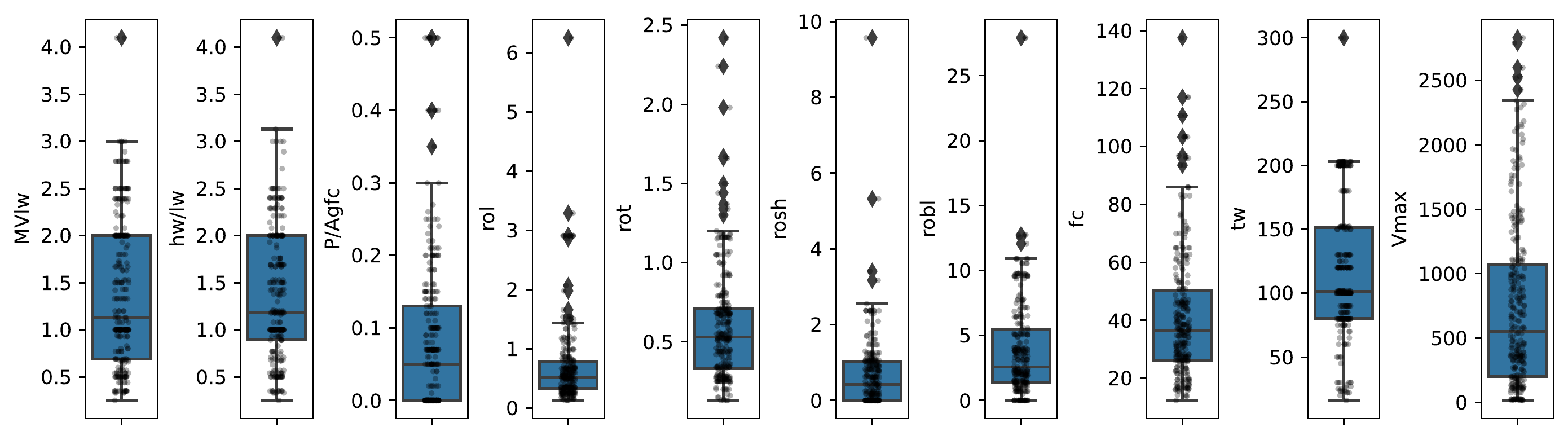}
	\caption{Distribution of the input variables in the database.}
	\label{fig:datadistribution}
\end{figure*}

The experimental data used in this research is a sub-assembly of the wall test database utilized in Deger and Taskin (2022) \cite{deger2022novel} with 30 additional data \cite{tokunaga2012experimental,hirosawa1975past}. As the main focus is to estimate the deformation capacity of walls governed by shear or shear-flexure interaction, walls that did not show so-called shear failure indications are excluded from the database, resulting in 286 specimens of use for this research. All specimens were tested under quasi-static cyclic loading, whereas none was retrofitted and re-tested. The database consists of wall design parameters, which are herein designated as the input variables of the machine learning problem, namely: wall geometry ($t_w$, $l_w$, $h_w$), shear span ratio ($M/Vl_w$), concrete compressive strength ($f_c$), longitudinal and transverse reinforcing ratios at web ($f_{yl}$, $f_{yt}$), longitudinal and transverse reinforcing ratios at boundary elements ($f_{ybl}$, $f_{ysh}$), axial load ratio ($P/A_gf_c$), shear demand (or strength) at the section ($V_{max}$), cross-section type (rectangular, barbell-shape, or flanged), curvature type (single or double). It is noted that single curvature and double curvature correspond to the end conditions of the specimen, i.e., cantilever and fixed-fixed, respectively. Distributions of the input variables are presented in Fig.\ref{fig:datadistribution} along with their box plots (shown in blue).

The output variable of the ML problem, the deformation capacity, is taken directly as the reported ultimate displacement prior to its failure if the specimen is tested until failure. Otherwise, it is assumed as the displacement corresponding to $0.8V_{max}$ as suggested by Park, 1989. It is noted that failure displacement was taken as the total wall top displacement and was not separated into shear and flexural deformation components.

\section{Explainable Boosting Machines}
%https://interpret.ml/docs/faq.html
Explainable Boosting Machines (EBM) is a state-of-the-art machine learning technique designed as accurately as random forests and boosted trees while also being simple to understand \cite{nori2019interpretml,nori2021accuracy}. The EBM delivers a complete explainable learning model that belongs to the family of Generalized Additive Models (GAMs) \cite{hastie1987generalized}:
\begin{equation}
    g(f(x_1,\ldots, x_n) ) = f_0 + f_1(x_1) + f_2(x_2) + \ldots, f_n(x_n) 
    \label{eq:gam}
\end{equation}
where $f_0$ is an intercept, and each $f_j$ is called a shape function, representing the individual effect of the $x_j$-th variable on the model output, $f(x_1,\ldots, x_n)$. The $g$ is utilized as a link function, adapting the model to different settings, e.g., identity function for regression and logistic function for classification. The intercept, $f_0$,  is calculated as the mean response of all the outputs. Because the shape functions are trained independently for each input variable, the model is additive (decomposable), allowing to separately analyze the effect of each input variable on the model output. The EBM is designed to improve the performance of the standard GAM while maintaining its interpretability.

Generalized Additive Models are more comprehensible than black-box models, but the analytical form of the shape functions is typically unknown, making it unsuitable for machine learning purposes. Although other analytical functions, such as splines or orthogonal polynomials, can be offered for defining shape functions, they are frequently less accurate when representing a nonlinear model \cite{mclean2014functional}.  The EBM uses shallow trees to construct the shape functions; therefore, it easily captures the nonlinearity of the data. Each input variable $(x_i)$ is modeled with ensemble trees such as bagging and gradient boosting. As a result, rather than employing the spline method, which is prevalent in traditional GAMs, the function associated with each input variable or interaction is produced from a vast set of shallow trees. 

\begin{figure*}[!h]
	\centering
	\includegraphics[width=4.8in]{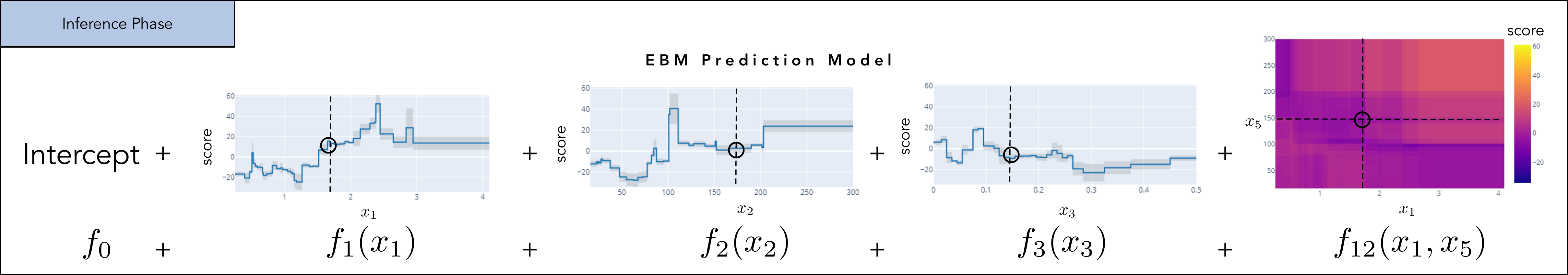}
	\caption{Inference phase of EBM.}
	\label{fig:ebm_model}
\end{figure*}

The EBM offers both local and global explanations of the learning model as each variable importance is estimated as the average absolute value of the predicted score. Moreover, each shape function can be visualized (algorithmically transparent); therefore, it is possible to observe the effects of the particular feature at certain intervals. In the inference phase, all the terms in Eq.\ref{eq:gam} are added up and passed through the link function to $g$ to compute the final prediction as shown in Fig. \ref{fig:ebm_model}.  In other words, individual predictions are generated using the shape functions, $f_i$, which act as a lookup table. 

To demonstrate which feature had the largest impact on the individual forecast, the contributions can be sorted and shown using the principles of additivity and modularity.  

The EBM's performance can be improved by including pairwise effects between variables in the model representation. For better performance, additional interactions can be incorporated; however, this may result in a more complex model with lower generalization performance due to the increased number of model parameters to be trained. The pairwise interactions are included in GA$^2$Ms \cite{lou2012intelligible}, which is a second-order additive model:
\begin{equation}
    g(E[y]) ) = f_0 + \sum_{i=1}^n f_i(x_i) + \sum_{i=1}^K f_k(x_{k_1},x_{k_2})
    \label{eq:gam2}
 \end{equation}
where $K$ pairs of features $(k_1, k_2)$ are chosen greedily (FAST algorithm) \cite{wood2008fast}. The pairwise interaction $f_{ij}(x_i,x_j)$ could be rendered as a heatmap on the two-dimensional $x_i$, $x_j$ - plane, still providing high intelligibility. Even though adding more interactions does not affect the model's explainability, the final prediction model may be less comprehensible due to a large number of interactions (less simulatability).

\section{Development of the Predictive Model}
\subsection{Overall Performance of EBM}
To assess whether the method compromises accuracy for the sake of interpretability, the performance of the EBM model is compared to three state-of-the-art black-box machine learning models, namely: XGBoost \cite{chen2016xgboost}, Gradient Boost  \cite{friedman2001greedy}, Random Forest \cite{breiman2001random}, and two glass-box models, namely Ridge Linear Regression \cite{hastie2009elements}, Decision Tree \cite{breiman2017classification}. All the implementations are carried out in a Python environment. For all ML models, the entire database, including all twelve input variables (ten variables from Fig.1 and two binary coded variables for curvature type and cross-section type), is randomly split into training and test datasets with a ratio of 90\% and 10\%, respectively.

\begin{figure}[!h]
	\centering
	\includegraphics[width=4in]{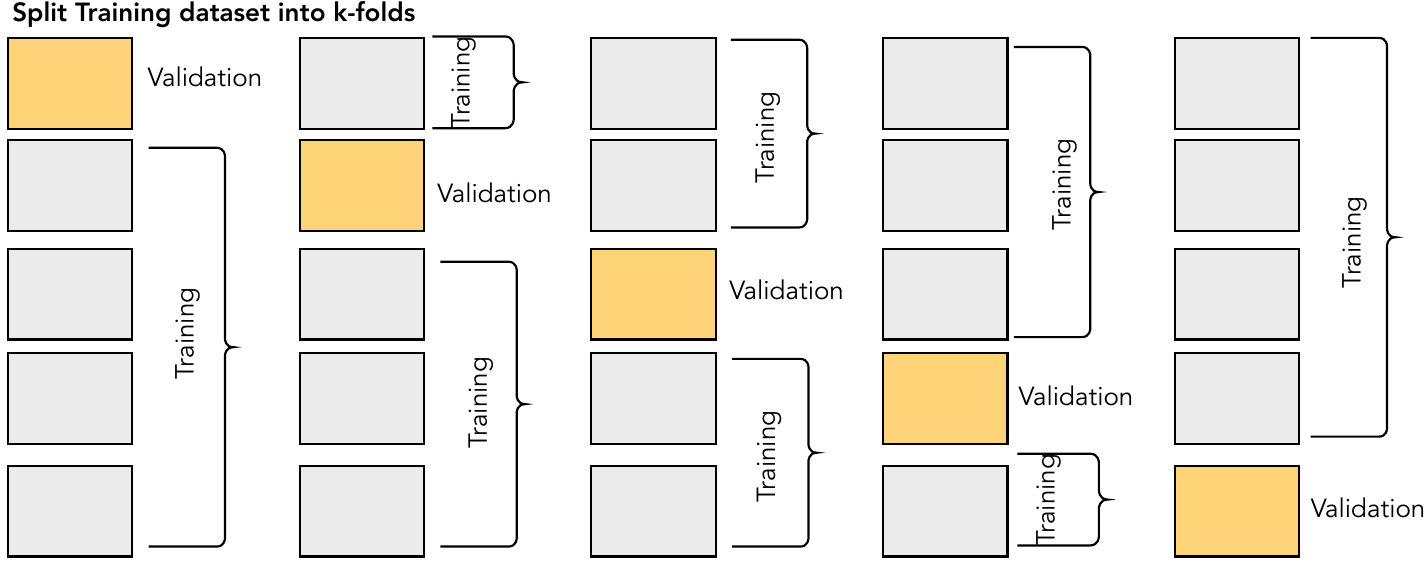}
	\caption{Illustration of k-fold cross-validation technique, where k is set to 5.}
	\label{fig:crossvalidation}
\end{figure}

Tunning of the hyperparameters, such as learning rate, number of leaves, number of interactions, and so on, typically affects the performance of the corresponding regression model. For hyperparameter tuning, a 10-fold cross-validation technique (Fig. \ref{fig:crossvalidation}) is used, wherein a subset of the data is kept as validation data, and the model's performance is evaluated using various hyperparameter settings on the validation set. This method prevents the tuning from overfitting the training dataset. 

For performance evaluations, the following three metrics are used over “unseen” (i.e., not used in the training process) test data sets of ten random train-test data splittings: coefficient of determination ($R^2$), relative error ($RE$), and prediction accuracy ($PA$), as given in Eqs. 3, 4, and 5, respectively.
\begin{eqnarray}
R^2 &=& 1-\frac{\sum_i (y_i -{\hat{y}}_i)^2}{\sum_i (y_i -{\bar{y}})^2} \\
RE  &=& \frac{\sum_{i} |\hat{y_i} - y_i|}{\sum_{i} |y_i|}|   \times 100\%   \\
PA &=& \sum_{i} \frac{y_i}{\hat{y_i}} 
\end{eqnarray}
where $y_i$, $\bar{y}$, $\hat{y_i}$, and $m$ refer to the actual output, the mean value of  $y_i$s, predicted output of corresponding regression model, and a number of samples in the test dataset, respectively.

\begin{figure*}[!t]
	\centering
	\includegraphics[width=5in]{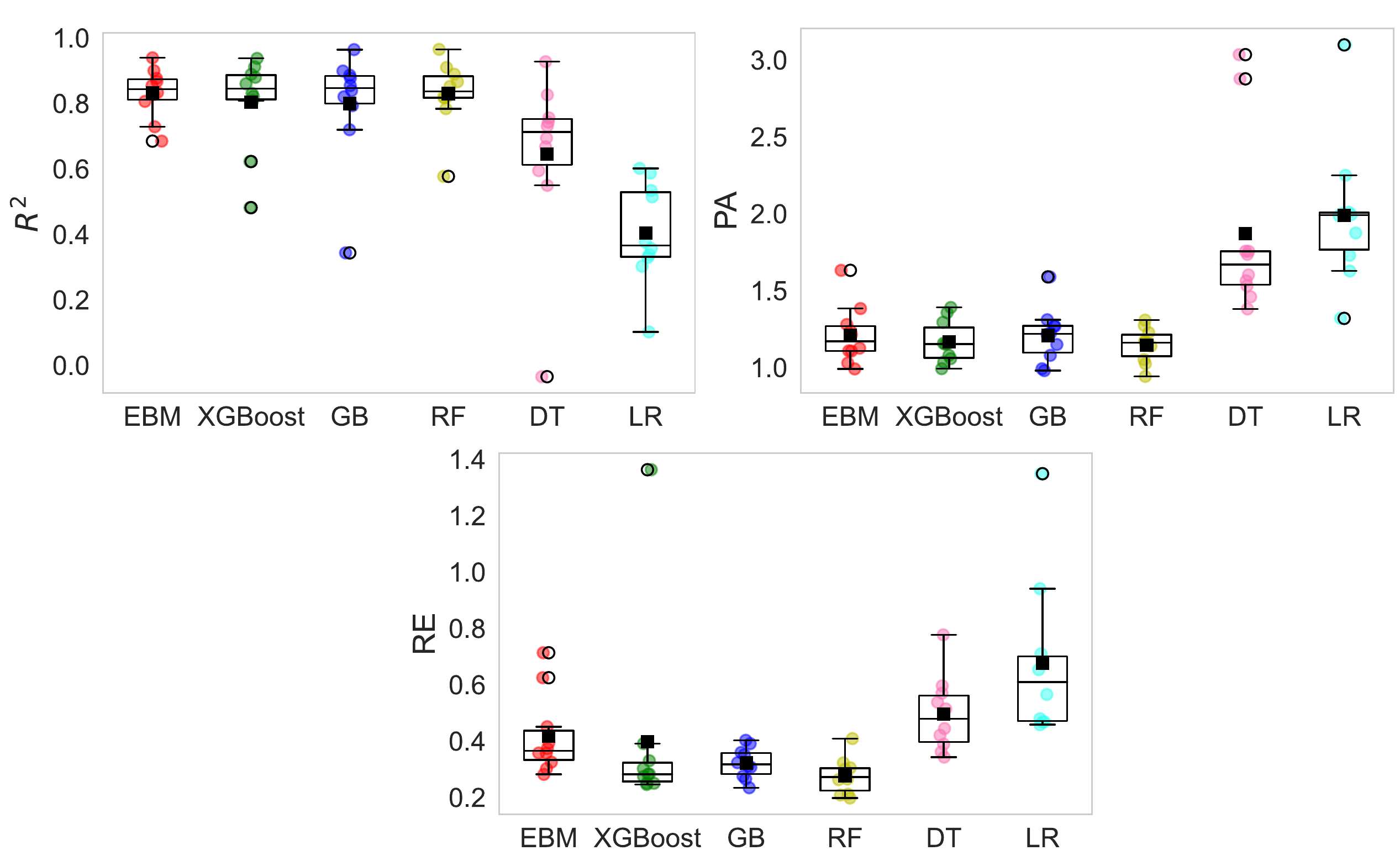}
	\caption{Comparison of performance scores of ML methods for test samples based on ten random train splittings.}
	\label{fig:boxplotsall}
\end{figure*}

Mean performance scores of the ML models are summarized in Table \ref{table:allresults}, along with their dispersion demonstrated in box plots in Fig. \ref{fig:boxplotsall}. The results indicate that EBM achieves comparable performance with its black-box counterparts, with a correlation of determination of $R^2=0.83$, a relative error of $0.41\%$, and a $PA=1.21$. As seen in Fig. \ref{fig:boxplotsall}, the low $R^2$, $RE$, and $PA$ deviations of EBM imply that reliable predictions can be achieved regardless of the selected train-test splitting and verify the model's robustness. Mean prediction accuracy (PA) shows around $20\%$ of overestimation for EBM and the black-box methods, suggesting that some input variables are potentially noisy. Compared to transparent models, the EBM outperforms both the Decision Tree (DT) and Ridge Linear Regression (RLR) across all three metrics, indicating that it is far superior to the traditional glass-box approaches. 

\begin{table}
\centering
\caption{Mean performance scores based on the test datasets over ten random splittings.}
\begin{tabular}{lllllll}
                 &              & \multicolumn{3}{l}{\textbf{Black-box}}  & \multicolumn{2}{l}{\textbf{Glass-box}} \\
                 & \textbf{EBM} & \textbf{XGBoost} & \textbf{GB} & \textbf{RF} & \textbf{RLR}       & \textbf{DT}       \\
                 \hline
\textbf{R2}      & 0.83         & 0.80             & 0.79        & 0.83        & 0.41               & 0.67              \\
\textbf{RE (\%)} & 0.41         & 0.40             & 0.32        & 0.28        & 0.67               & 0.47              \\
\textbf{PA}      & 1.21         & 1.17             & 1.20        & 1.15        & 2.0                & 1.9              \\
\hline
\end{tabular}
\label{table:allresults}
\end{table}

The most remarkable advantage of the EBM method over the others is that it provides full explainability without sacrificing accuracy. Unlike other methods, EBM enables the user to understand how the prediction is made and which parameters are essential in the decision-making process. Therefore, the EBM method is selected as the baseline algorithm for the rest of the analysis to propose a  prediction model for estimating the deformation capacity based on the following criteria: developing a model with fewer input variables (high simulatability), achieving high accuracy, and ensuring physical consistency.

\subsection{The Proposed EBM-based Predictive Model}

\begin{figure}[!h]
	\centering
	\includegraphics[width=4.5in]{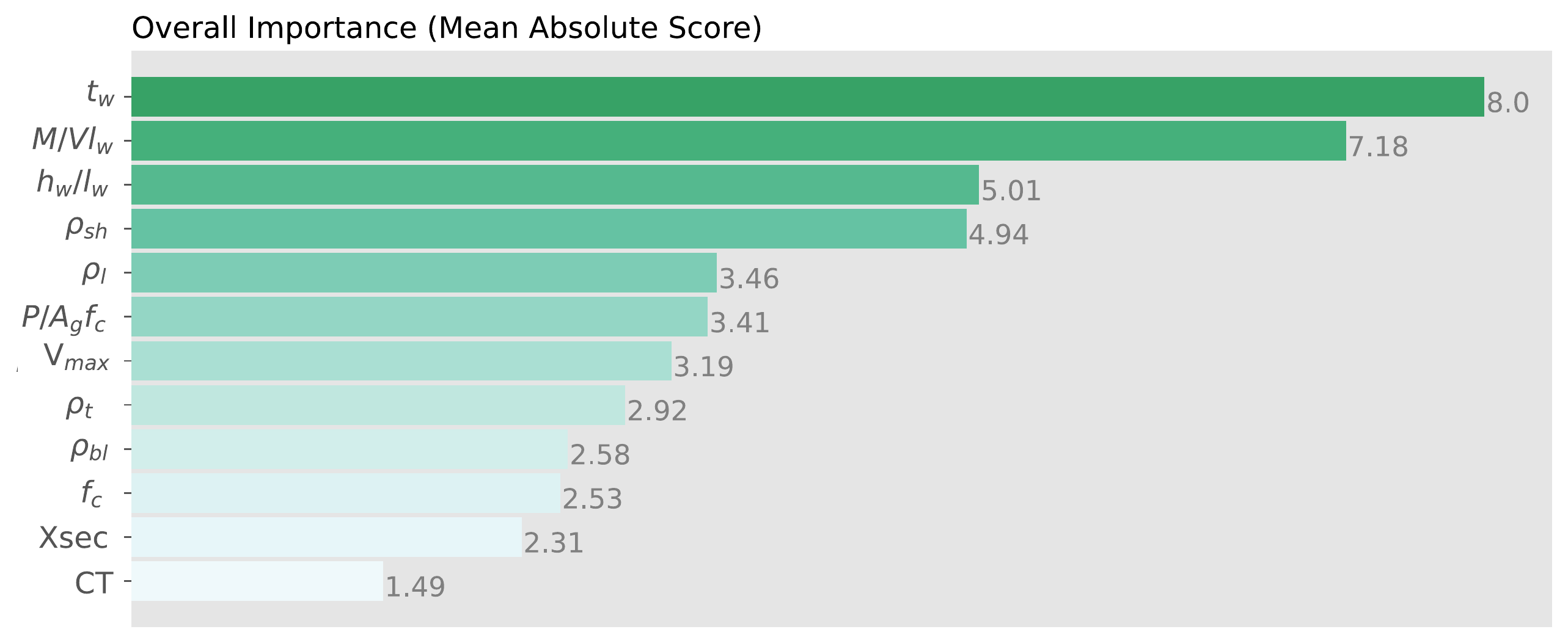}
	\caption{EBM Global interpretation for twelve features included}
	\label{fig:allfeaturesScores}
\end{figure}

The importance of the wall properties in predicting the deformation capacity is evaluated based on additive term contributions visualized in Fig.\ref{fig:allfeaturesScores}. Results reveal that $t_w$ and $M/Vl_w$ (or $h_w$/$l_w$) have the greatest impact on individual predictions. This is consistent with the mechanics of the behavior as walls with smaller thickness are shown to be more susceptible to lateral stiffness degradation due to concrete spalling, leading to a failure caused by lateral instabilities or out-of-plane buckling \cite{vallenas1979hysteric,oesterle1976earthquake}. The shear span ratio (or aspect ratio), on the other hand, both have a significant impact on deformation capacity as the higher the shear span ratio gets, the slender the wall is, and the higher deformations it typically can reach prior to its failure. The least important wall parameters, on the other hand, are identified as curvature type, cross-section type, and concrete compressive strength. 

Another critical aspect considered in this study is to develop the predictive model with as few input variables as possible. With that, the computational workload is aimed to be reduced, and a more practical and interpretable model is proposed for potential users. To achieve this, knowledge-based-selected combinations of four-to-five features are exhaustively evaluated to reach performance scores as high as when twelve features are included. 

EBM can achieve similar performance scores using four features: $M/Vl_w$, $P/A_gf_c$, $t_w$, and $V_{max}$. Including additional features (e.g., $\rho_l$, $\rho_{bl}$, $\rho_{sh}$) deemed impactful  by EBM as well as experimental results \cite{tasnimi2000strength,hube2014seismic} has only a modest effect on the overall performance. The performances of other methods are close to their benchmark model (including twelve features), whereas the glass-box methods are affected by the reduction of input size and show much lower performances. The mean $R^2$ drops to $0.33$ for the Ridge Linear Regression, imparting that the input-output relation is not linear. 

\begin{figure}[!h]
	\centering
	\includegraphics[width=3.5in]{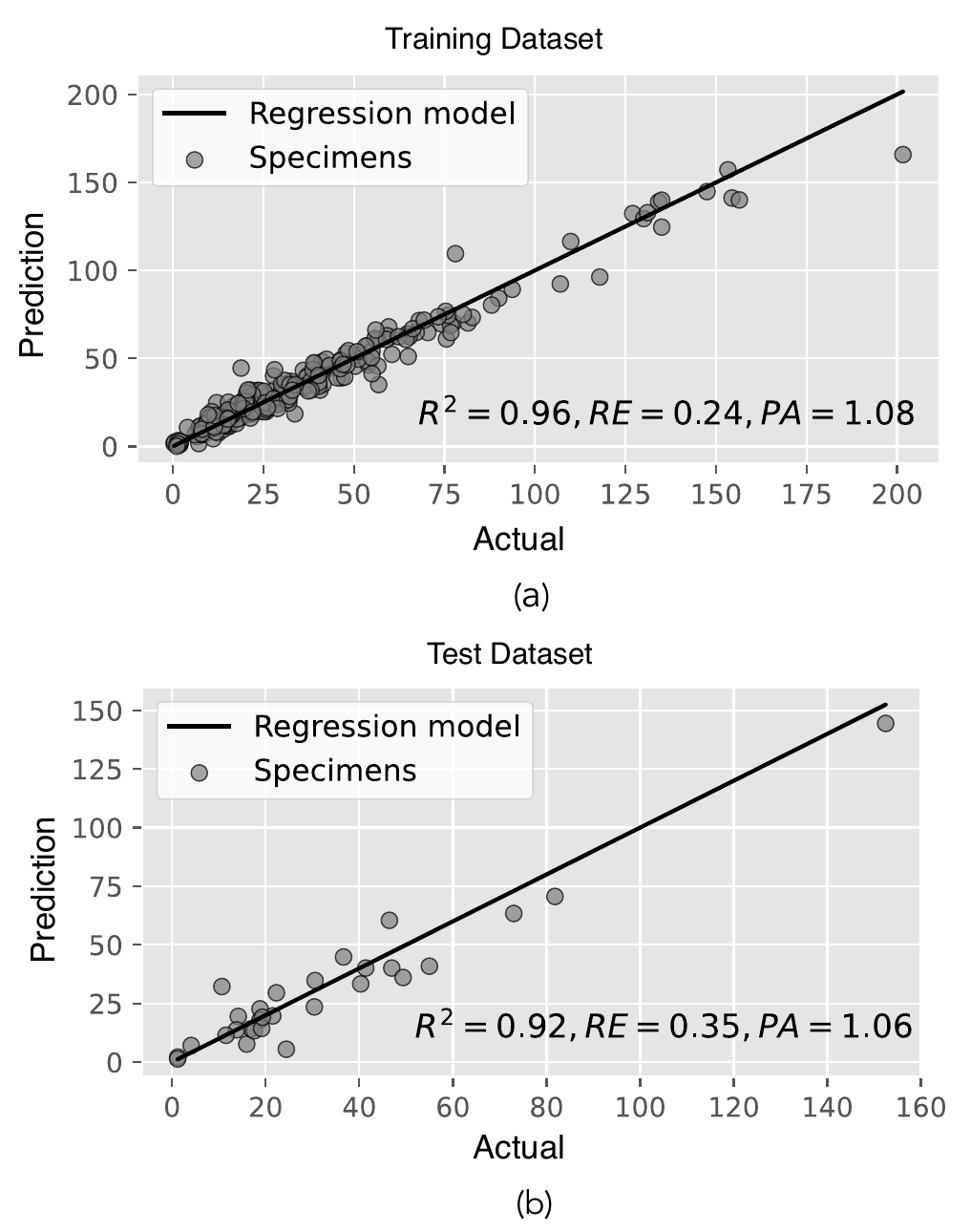}
	\caption{Correlations of the model outputs with the actual values for (a) training and (b) test datasets}
	\label{fig:corrPlotsAll}
\end{figure}

The proposed EBM-based predictive model is selected to achieve the highest $R^2$ with a prediction accuracy as close to 1.0 as possible. The correlation plots are presented in Fig. \ref{fig:corrPlotsAll} for training and test data sets, where scattered data are concentrated along the $y=x$ line, demonstrating that the proposed model can make accurate predictions. It should be noted that the distribution of the residuals is concentrated around zero.

\begin{figure*}[!h]
	\centering
	\includegraphics[width=5.5in]{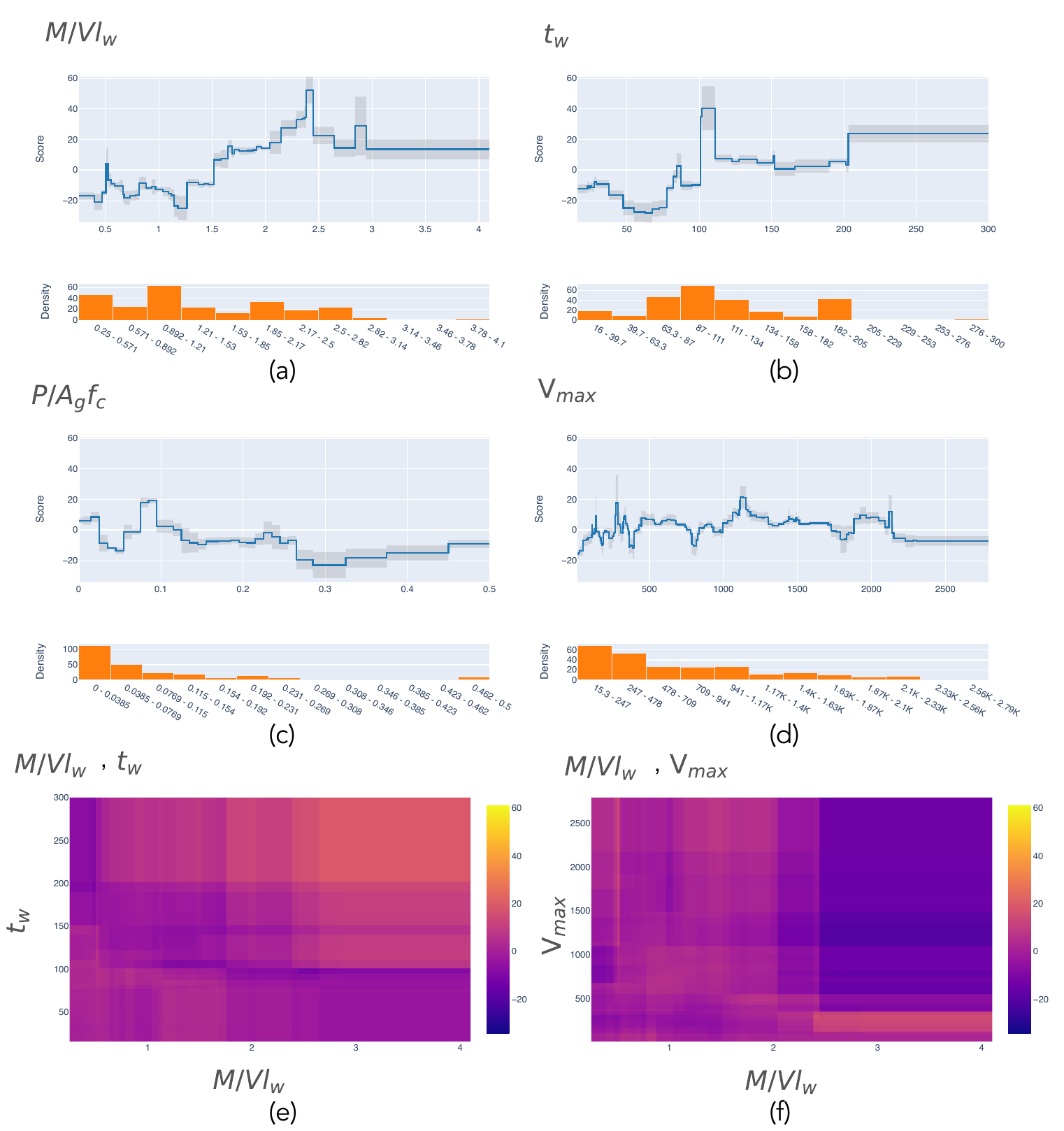}
	\caption{EBM shape functions (a-d) and pairwise interaction plots (e-f) for the proposed model. Note that the intercept $f_0$= 35.528.}
	\label{fig:proposed_ebm}
\end{figure*}

As discussed above, the proposed model is an additive model in which each relevant feature is designated a quantitative term contribution. The EBM allows the user to explore the contribution of each feature to the model by plotting their shape functions (Fig.\ref{fig:proposed_ebm}a-d). As discussed above, the EBM method employs multiple decision-tree learning models; therefore, inclines and declines are undertaken with jump-looking piece-wise constant functions (versus smooth curves). The values, called scores, are read from these functions, and those from heat maps (Fig.\ref{fig:proposed_ebm}e-f) representing pairwise interactions (i.e., between two features) are summed up to calculate the prediction. The gray zones along the shape functions designate error bars that indicate the model's uncertainty and data sensitivity. This typically occurs in cases of sparsity or the presence of outliers within the associated region.

% {\color{blue}
%The error bars are rough estimates of the uncertainty of the model in each region of the feature space. A large error bar means that the learned function may have changed substantially with minor changes to the training data, and indicates that the interpretation of the model in that region should be treated with more skepticisim.

%The size of the error bar is typically determined by two factors – the amount of training data in that region of the feature space, and the inherent uncertainty of the learned model. For example, consider this graph of the learned “Age” feature from the Adult income dataset:}
\begin{figure*}[!t]
	\centering
	\includegraphics[width=6in]{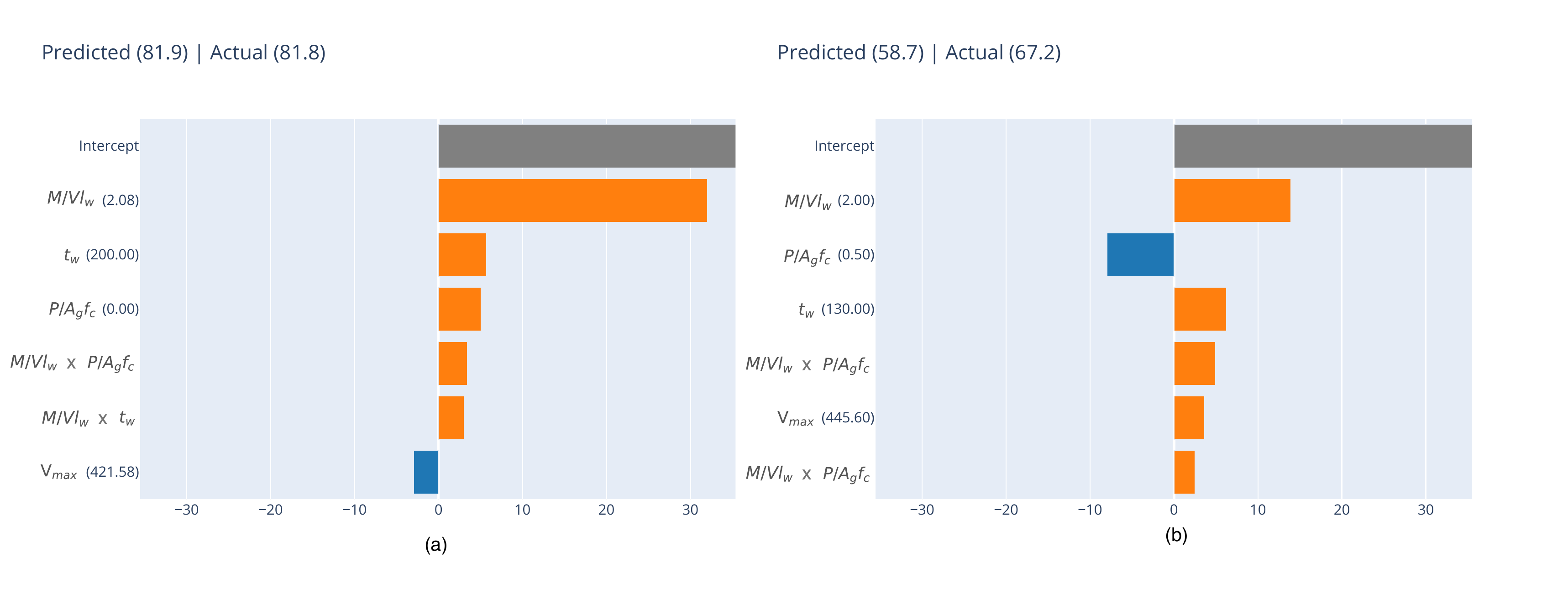}
	\caption{Variable contribution estimates for (a) well-predicted, (b) averagely-predicted samples.}
	\label{fig:sample}
\end{figure*}
The shape functions in Fig.\ref{fig:proposed_ebm} also indicate their correlations with the output in a graphical representation. For example, nonlinear patterns that can not be observed in linear approaches can be easily interpreted \cite{zschech2022gam}, which provides new insights to broaden existing experimental-based knowledge. For example, the shear demand $V_{max}$ (Fig.\ref{fig:proposed_ebm}d) reduces ductility; thus deformation capacity, as demonstrated by experimental results \cite{corley1981structural} and suggested by ASCE 41-17 acceptance criteria. Yet, a highly nonlinear pattern is observed when relevant experimental data are gathered \cite{aladsani2022explainable,deger2019empirical}. This nonlinearity can be observed in the shape function suggested by the proposed method. Other input variables ($t_w, P/A_g f_c, M/Vl_w$), on the other hand, demonstrate an almost-linear trend. The interpretation of EBM for these variables is consistent with experimental results in the literature, such that $M/Vl_w$  (Fig.\ref{fig:proposed_ebm}a) and $t_w$ (Fig.\ref{fig:proposed_ebm}b) have a positive impact, as discussed above, whereas $P/A_gf_c$ (Fig.\ref{fig:proposed_ebm}c) has an adverse influence \cite{lefas1990behavior,farvashany2008strength}. The reason for $M/Vl_w$ and $t_w$ (Fig.\ref{fig:proposed_ebm}b) suggesting an inverse effect up to a certain point ($M/Vl_w \approx 1.2$, $t_w \approx 60$ cm,  $P/A_gf_c \approx 0.08$) is because the model has an intercept value ($f_0$, Eq.\ref{eq:gam2}) and specimens with smaller deformation capacities ($f_0$ less than $35.528$) are predicted adding up negative values. The unexpected jumps in $t_w$ are likely because there is an abrupt accumulation of data at $t_w = 100$ mm and $t_w = 200$ mm (64 and 44 specimens, respectively), which probably causes difficulty in decision making. 

It is noted that the EBM method offers controllability over the structure of the model proposed by, for instance, modifying the number of pairwise interactions. This allows the method to suggest more than one model for the same input-output configuration for a particular train-test dataset. Reducing the number of interactions brings simplicity to the model; however, it typically loses accuracy as EBM relies on its automatically-determined interactions in the decision-making process. Given this trade-off, the number of interactions is set to two for the proposed model. 

\subsection{Sample-Based Explanation}
This section presents the prediction of deformation capacity for two example specimens using the proposed EBM-based predicted model. One specimen is predicted with excellent accuracy (almost zero error; Fig.\ref{fig:sample}a), whereas the other is predicted with around 15\% error (Fig.\ref{fig:sample}b).

Variable contribution estimates for each specimen are presented such that the intercept is constant and shown in gray, the additive terms with positive impact are marked in orange, and additive terms decreasing the output are shown in blue. Each contribution estimate is extracted from the shape functions and two-dimensional heat maps (Fig.\ref{fig:proposed_ebm}) based on the input values of a specific specimen. Overall, the model is consistent with physical knowledge, except $V_{max}$ has an unexpected positive impact on the output for the relatively worse prediction (Fig.\ref{fig:sample}b). This is an excellent advantage of EBM; that is, the user can prudently understand how the prediction is made for a new sample and develop confidence in the predictive model (versus blind acceptation in black-box models).

\subsection{Comparisons with Current Code Provisions }
ASCE 41-17 and ACI 369-17 \cite{aci369} provide recommended deformation capacities for nonlinear modeling purposes, where shear walls are classified into the following two categories based on their aspect ratio: shear-controlled ($h_w/l_w > 1.5$) and flexure-controlled ($h_w/l_w > 3.0$). The deformation capacity of shear-controlled walls is identified as drift ratio such that $\Delta_u/h_w = 1.0$ if the wall axial load level is greater than 0.5 and $\Delta_u/h_w = 2.0$ otherwise. 

\begin{figure}[!h]
	\centering
	\includegraphics[width=3.5in]{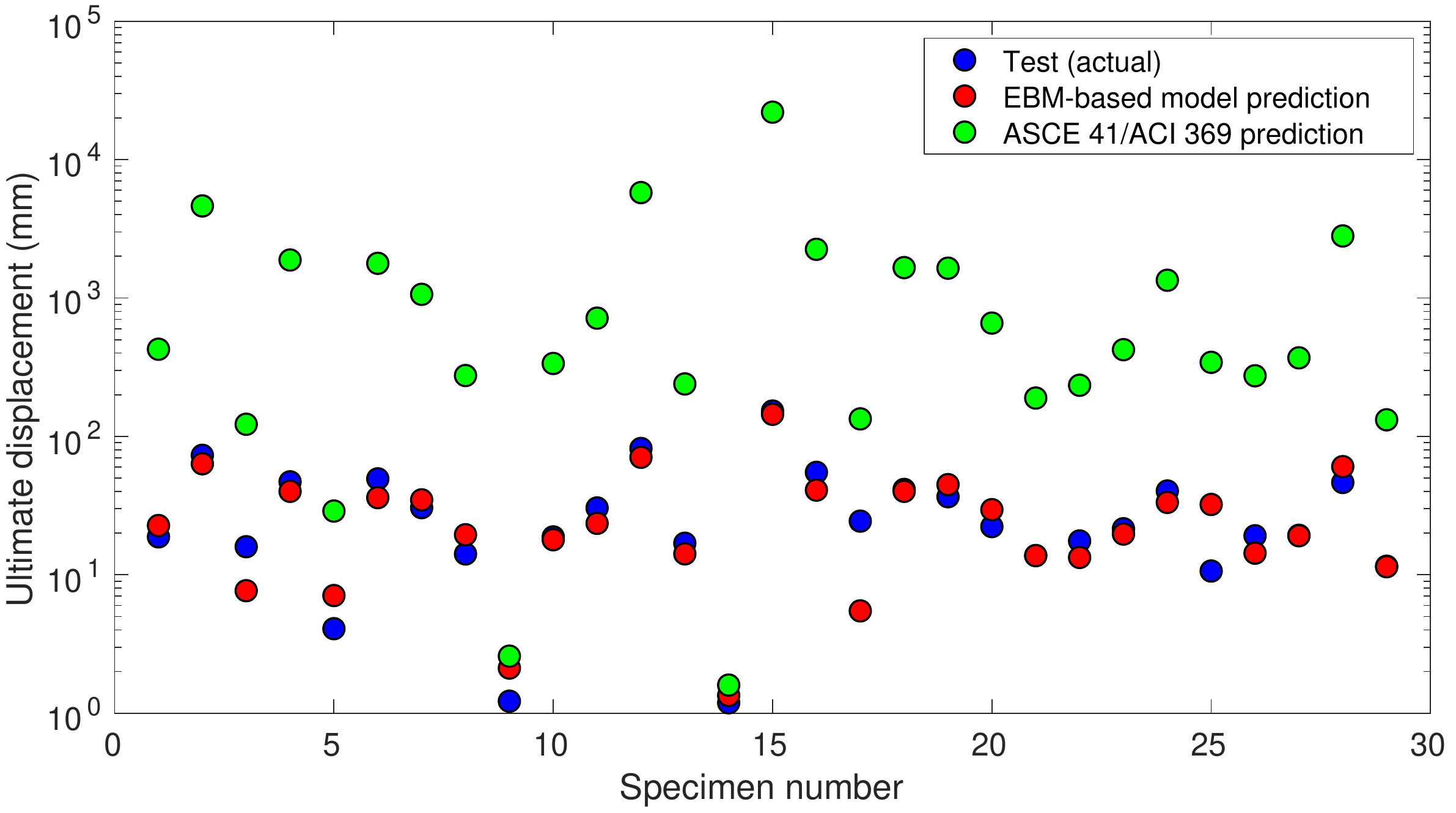}
	\caption{Comparisons of EBM-based model predictions with code provisions.}
	\label{fig:comparison}
\end{figure}

Deformation capacity predictions based on the proposed EBM model are compared to ASCE 41-17/ACI 369-17 provisions in Fig. \ref{fig:comparison}. Predicted-to-actual ratios are $1.06\pm0.49$ and $6.42\pm3.17$ for EBM-based model and code predictions, respectively. The results imply that traditional approaches may lead to the overestimation of deformation capacities and cause unsafe assessments.

\section{Conclusions}
A fully transparent predictive model is developed to estimate the deformation capacity of reinforced concrete shear walls that are failed in pure shear or shear-flexure interaction. To achieve this, a state-of-the-art machine learning method, Explainable Boosting Machines (EBM), designed as accurately as random forests and boosted trees, is utilized. The EBM provides an additive model such that each relevant feature is designated a quantitative term contribution. The input-output configuration of the model is designated as the shear wall design properties (e.g., wall geometry, axial load ratio) and ultimate wall displacement, respectively. The conclusions derived from this study are summarized as follows:
\begin{itemize}
    \item The importance of the wall properties in predicting the deformation capacity is evaluated based on additive term contributions. $t_w$ and $M/Vl_w$ (or $h_w$/$l_w$) have the greatest effect on individual predictions, whereas the least relevant ones are identified as curvature type, cross-section type, and concrete compressive strength.

\item	Compared to three black-box models (XGBoost, Gradient Boost, Random Forest), the EBM achieves similar or better performance in terms of correlation of determination ($R^2$), relative error ($RE$), and prediction accuracy ($PA$; the ratio of predicted to the actual value). The EBM achieves a mean $R^2$ of 0.83 and a mean $RE$ of 0.41\% using twelve input variables based on ten random train-test splittings. 
\item Compared to two glass-box methods (Decision Tree (DT) and Ridge Linear Regression (RLR)), the EBM outperforms both methods across all three metrics. 
\item The dispersion of performance metrics of EBM is small, implying that the model is robust and the performance is relatively less data-dependent.
\item Compared to the developed model when  all the available features are used, the EBM achieves competitive performance scores using only four input variables: $M/Vl_w$, $P/A_gf_c$, $t_w$, and $V_{max}$. Using these four features, the proposed EBM-based model achieves $R^2$ of 0.92 and $PA$ of 1.05 based on the test dataset. Using fewer variables ensures that the model is less simulatable, more practical, more comprehensible, and reduces the computational cost.
\item It is important to note that the decision-making process developed by the proposed EBM-based model has overall consistency with scientific knowledge despite several exceptions detected in sample-based inferences. This is an excellent advantage of the proposed model; that is, the user can assess and evaluate the prediction process before developing confidence in the result (versus blindly accepting as in black-box models).
\item This model delivers exact intelligibility, i.e., there is no need to use local explanation methods (e.g., SHAP, LIME) to interpret the learning model, which obviates the uncertainties associated with their approximations. 
\end{itemize}

The proposed EBM-based model is valuable in that it is simultaneously accurate, explainable, and consistent with scientific knowledge. The EBM's ability to provide interpretable and transparent results would allow engineers to better understand the factors that affect the deformation capacity of non-ductile RC shear walls and make informed design decisions. The use of the EBM to estimate deformation capacity would improve the reliability and efficiency of structural analysis and design processes, leading to safer and more cost-effective buildings.

\newpage
\clearpage

%%%%%%%%%%%%%%%%%%%%%%%
\bibliographystyle{unsrt}
 \bibliography{main-bib}

\end{document}